%% file: lrec2022-example.tex
\documentclass[10pt, a4paper]{article}
\usepackage{lrec2022} % this is the new LREC2022 Style
\usepackage{multibib}
\newcites{languageresource}{Language Resources}
\usepackage{graphicx}
\usepackage{tabularx}
\usepackage{soul}

\usepackage{times}
\usepackage{latexsym}

\usepackage[T1]{fontenc}
\usepackage[utf8]{inputenc}

\usepackage{microtype}

\usepackage{graphicx}
\usepackage{newunicodechar}
\usepackage[T1]{fontenc}
\usepackage{textcomp}
\usepackage{lmodern}
\usepackage{amsmath}
\usepackage{amssymb}
\usepackage{multirow}
\usepackage{booktabs}
\usepackage{float}

\usepackage{tikz}
\usetikzlibrary{
  shapes,
  arrows,
  arrows.meta,
  shapes.geometric,
  calc,
  positioning,
  backgrounds,
  decorations.pathreplacing,
  matrix,
  math
}

\newcommand{\placeholder}{\ensuremath{\square}}
\newcommand{\sentenceexample}{He was a son of David and Mary M Anderson}
\newcommand{\highlightedexample}{\textit{He} and \textit{Mary M Anderson}}

\usepackage{titlesec}
\titleformat{\section}{\normalfont\large\bfseries\center}{\thesection.}{1em}{}
\titleformat{\subsection}{\normalfont\SmallTitleFont\bfseries\raggedright}{\thesubsection.}{1em}{}
\titleformat{\subsubsection}{\normalfont\normalsize\bfseries\raggedright}{\thesubsubsection.}{1em}{}
\renewcommand\thesection{\arabic{section}}
\renewcommand\thesubsection{\thesection.\arabic{subsection}}
\renewcommand\thesubsubsection{\thesubsection.\arabic{subsubsection}}
\usepackage{epstopdf}
\usepackage[utf8]{inputenc}

\usepackage{hyperref}
\usepackage{xstring}

\usepackage{color}

\title{From Examples to Rules: Neural Guided Rule Synthesis for Information Extraction}

\name{\begin{tabular}{c}Robert Vacareanu, Marco A. Valenzuela-Esca\'rcega, George C. G. Barbosa, \\
Rebecca Sharp, Mihai Surdeanu\end{tabular}}

\address{University of Arizona \\
         Tucson, AZ, USA \\
         \{rvacareanu, gcgbarbosa, msurdeanu\}@email.arizona.edu\\\{marcov, bsharpataz\}@gmail.com\\}
\abstract{
While deep learning approaches to information extraction have had many successes,
they can be difficult to augment or maintain as needs shift.
Rule-based methods, on the other hand, can be more easily modified.
However, crafting rules requires expertise in linguistics and the domain
of interest, making it infeasible for most users.
Here we attempt to combine 
the advantages of these two directions while mitigating their drawbacks.
We adapt recent advances from the adjacent field of program synthesis to 
information extraction, synthesizing rules from provided examples.
We use a transformer-based architecture to guide an enumerative search, and show that
this reduces the number of steps that need to be explored before a rule is found.
    Further, we show that %\textit{without training on the specific domain}, % too strong...
    {\textit{without training the synthesis algorithm on the specific domain,}}
    our synthesized rules
achieve state-of-the-art performance on the 1-shot scenario of a task that focuses on few-shot learning for relation classification, and competitive performance in the 5-shot scenario. 
 \\ \newline \Keywords{rule-based information extraction, rule synthesis} }

\begin{document}

\maketitleabstract

\input{tex/introduction}

\input{tex/related_work}

\input{tex/problem}

\input{tex/method}

\input{tex/experiments}

\input{tex/conclusion}

% \citelanguageresource{EMILLE}
% \cite{manning2015computational}

% \section*{Appendix: How to Produce the \texttt{.pdf}}

% In order to generate a PDF file out of the LaTeX file herein, when citing language resources, the following steps need to be performed:

% \begin{itemize}
%     \item{Compile the \texttt{.tex} file once}
%     \item{Invoke \texttt{bibtex} on the eponymous \texttt{.aux} file}
%  %   \item{Invoke \texttt{bibtex} on the \texttt{languageresources.aux} file}
%     \item{Compile the \texttt{.tex} file twice}
% \end{itemize}

% \nocite{*}
\section{Bibliographical References}\label{reference}
%\label{main:ref}

\bibliographystyle{lrec2022-bib}
\bibliography{lrec2022-example}

\section{Language Resource References}
\label{lr:ref}
\bibliographystylelanguageresource{lrec2022-bib}
\bibliographylanguageresource{languageresource}

\end{document}

%% file: tex/introduction.tex
\section{Introduction}

The ``deep learning tsunami'' that ``hit'' natural language processing (NLP)~\cite{manning2015computational} has brought tremendous improvements in performance to most NLP applications. 
However, these benefits do not come for free. One drawback of deep learning is its opacity, which limits the ability for users to make incremental improvements to deployed systems.  In particular, the entanglement of deep learning approaches means that ``changing one thing changes everything''~\cite{sculley2015hidden}. 
% ms: I think the cleanest solution is to just take this sentence out:
%Consequently, many domain-specific NLP applications, especially in industry, continue to focus on rule-based approaches~\cite{chiticariu2013rule,wang2018clinical,ValenzuelaEscarcega2018LargescaleAR}.

In contrast, rule-based approaches are  much more amenable to incremental improvements as each individual rule can be interpreted explicitly and unambiguously.
This is critical for systems which will be deployed and maintained for long periods of time.
Further, because rules encode expert knowledge, experts can write them without first curating a large number of examples.

%In contrast, rule-based methods are interpretable~\cite{chiticariu2013rule}.

However, an  important drawback to rule-based  approaches is that rule development is time consuming, and requires expertise  in  both the domain at hand and in linguistics. 
To mitigate this limitation, many directions before the ``deep learning tsunami'' focused on rule learning from examples~\cite[{\em inter alia}]{yarowsky1995unsupervised,riloff1996automatically,collins1999unsupervised,abney2002bootstrapping,mcintosh2010unsupervised,gupta2014improved}. 

%These approaches, however, have seldom been revisited since deep learning revolutionized the NLP field. % ms : this is a dangerous statement
Here we propose a novel method for rule synthesis from examples that combines the strengths of deep learning with the advantages of rule-based methods.  
By utilizing a self-supervised pre-trained transformer, we minimize the number of examples needed from the expert.
By generating human-readable rules, the resulting grammar can be adjusted or extended incrementally as needs shift.

Our ability to generate rules from limited examples is key to our approach as this mimics a real-world setting, where the cost (both in terms of time and money) of annotating thousands of examples for training supervised approaches is a barrier for many.  
Accordingly, we evaluate our methods in a \textit{few-shot} framework, to simulate a user providing a small number of examples from which to generalize.

The key contributions of this paper are:

{\flushleft {\bf (1)}} To our knowledge, we are the first to propose methods inspired from program synthesis for rule learning in IE. 
Our method includes a contextualized neural component that scores each intermediate state in the rule synthesis process based on its likelihood to lead to a good rule, with backtracking facilitated with the Branch and Bound algorithm~\cite{bnb}.
%Our approach is compositional and self-supervised, and thus, domain agnostic.
All data and code needed to replicate are open-source and publicly available.\footnote{The code will be made available at this URL: \url{https://github.com/clulab/releases/tree/master/lrec2022-odinsynth}.}

{\flushleft {\bf (2)}} In an intrinsic evaluation, we show that our neural-guided rule synthesis reduces the number of 
search steps considerably compared to a synthesis method using static state scores. 
Importantly, similar to language models, our neural guiding function doesn't use in-domain training data, which means that it works without training or fine-tuning on any IE domain.

{\flushleft {\bf (3)}} 
% We demonstrate that our approach obtains competitive performance on an extrinsic IE task, despite the fact that % our evaluation is ``zero shot,'' i.e., % ms: seems too strong since we do look at the training data
%our contextualized state scorer is never exposed to annotations or  texts from the IE task. 
In an extrinsic evaluation we demonstrate the validity of our rule synthesis approach.  In particular, we evaluate on the Few-Shot variant of the TACRED relation extraction task \citelanguageresource{zhang2017tacred,Sabo2021RevisitingFR}, and show that 
our approach considerably outperforms the state-of-the-art BERT model on the harder 1-shot task by $3\%$ F1 points.

%Additionally, our method is interpretable.}

% a rule-based method that relies on patterns manually-crafted for the task \cite{zhang2017tacred}, as well as sequence-to-sequence methods trained on the same data as our method. 
% \todo{Mihai: we are clustering on tacred}

% \begin{figure}[]
%     \centering
%     \includegraphics[width=0.9\columnwidth]{figures/sentence_example_new2.png}
%     \caption{\footnotesize{
%             Example of input/output for our rule synthesis method. Our approach learns to generate individual rules that match a set of examples (or a {\em specification}), where each example is a highlighted span of text in a sentence. Note that this solution is not unique.}
%     }
%     \label{fig:sentence}
% \vspace{-4mm}
% \end{figure}

%% file: tex/related_work.tex
\section{Related Work}

Our approach lies at the intersection of program synthesis and rule learning for IE.

{\flushleft {\bf Program synthesis:}} There has been a large body of work on methods for program synthesis, with a general focus on automatically
generating code \cite{flashfill,lee2016synthesizing,balog2016deepcoder,gulwani2017program}.
% survey -- gulwani2017program
In general, program synthesis requires a program space (the domain specific language),
the user intent (specification), and a search algorithm, from which it produces an
executable program.
%
%
% modern program synthesis:
%  input:
%  - program space (DSL and language/grammar)
%  - user intent (examples, natural language)
%  - search algorithm -- enumerative/deductive search, nn (seq2seq)
%  output:
%  - program
% Some early examples of successful program synthesis include FlashFill \todo{cite},
% a component of Excel that syntesizes a small program to
% manipulates strings in certain spreadsheet cells to create a new string.
%
One popular method for program synthesis is \textit{program by example} \cite{cypher1993watch,lieberman2001your,6481005}, which learns from specifications in the form of (input, output) pairs. % as examples to the
% search algorithm, 
These methods perform a deductive search over the program space
until a successful program is produced. Our rule synthesis method is situated within this framework.  
Importantly, as the program space becomes larger, this deductive search is intractable without intervention.
There are different forms of intervention, including heuristic pruning \cite{lee2016synthesizing}, 
and usage of a neural guiding function \cite{balog2016deepcoder,kalyan2018neural}.
In our work we make use of both, 
% \todo{cite} use heuristic {\em pruning} to reduce the search space.  \todo{cite} instead \textit{guide} the search by using a neural network to make an initial prediction of the likelihood that a given node or operation in the DSL will be useful for the final program.  \todo{cite} extend this idea, predicting at each step in the search.  Here, we incorporate both pruning and guiding, 
using a transformer network to guide our search as well as custom heuristics to prune whole branches from the search tree.

Beyond simply using neural networks to guide, there have been efforts to
generate the final program using a neural sequence-to-sequence model, e.g. \cite{yin2017syntactic}.
With these approaches, execution guidance is typically used
to ensure that the generated program is valid in the DSL
\cite{wang2018robust}.
This is not needed in our approach, since we make predictions over the possible next states 
as determined by the DSL grammar, which ensuring that every generated program is valid.%\todo{and only advance on valid continuations?}, obviating the need for this.

%\todo{cite} propose an approach that intentionally maintains elements from the input in the
%output, e.g., as with table names.
% PointerSQl - maintain elements from the input to the ouput unchanged.
%We do not make use of this in our implementation, but  this could potentially be applied in future work to allow the user to insist on portions of the specification appearing in the resulting rule.

% deductive -- e.g., flash meta
% give examples, seacrch for programs
% if you need to explore whole space, too slow
% prune with heuristics from regex paper.

% deep coder, has NN predicts what operations from DSL you'll need
% encode examples, produce probabilities over the DSL nodes or operations
% thus, you can prioritize over the search space, pruning elements you're
% unlikely to use.
% kalyan et al. instead use neural-guided deductive search to predict at each step.
% We are like the latter.

Aside from program by example, other approaches use different forms of specifications.
For example, \cite{dong2018coarse} and \cite{hwang2019comprehensive} generate programs from a natural language description of the desired behavior. %and \todo{cite} who use
%sequential encoder decoders that first transform natural language to an intermediate \textit{sketch}, and then tranform the sketch to an executable program.
However, for our work, we choose to focus on examples, as we feel it is a more intuitive interface; it can be very difficult to describe explicitly the desired behavior of an IE rule, especially when the user does not have experience with linguistic structures.

% Seq2Seq,  execution guidance (wang et al, 2018) to constrain the output to be syntactically valid in the DSL.

% coarse to fine seq2seq (natlang, generate sketch that is not executable but has elements needed, then another seq2seq that goes from sketch to program.)
% generate the sketches with heuristics, supervise both.
% we don't have traingin data with natural language.

%%% CITE THESE:
%
%Some papers:
%Synthesizing  Regular  Expressions  from  Examplesfor  Introductory  Automata Assignments \\
%A Syntactic Neural Model for General-Purpose Code Generation \\
%A Comprehensive Exploration on WikiSQLwith Table-Aware Word Contextualization \\
%
%
%Tangentially: \\
%Simulating SQL Injection Vulnerability ExploitationUsing Q-Learning Reinforcement Learning Agents \\
%pix2code: Generating Code from a Graphical User Interface Screenshot
%
%

{\flushleft {\bf Rule learning for information extraction:}} 
%IE also has a rich body of literature, too large to be properly summarized here. 
At a high-level, IE approaches fall in one of three camps: (a) methods that rely on rules or patterns (either manually crafted, learned, or extracted using a shortest path through the syntactic graph \cite{yarowsky1995unsupervised,riloff1996automatically,collins1999unsupervised,abney2002bootstrapping,mcintosh2010unsupervised,chiticariu2010systemt,gupta2014improved,shlain-etal-2020-syntactic}, (b) approaches that use ``traditional'' machine learning \cite{mintz2009distant,riedel2010modeling,hoffmann2011knowledge,surdeanu2012multi}, and (c) neural approaches \cite{zeng2015distant,lin2016neural,zhang2018graph,guo2019attention}.  
Our approach is closest to the first camp, in the sense that we output rules, but we use state-of-the-art methods from the last camp 
(such as transformer networks) to generate these rules. 

% ms: removed for space. we said it in the intro
%As such, we aim to take advantage of the high performance of deep learning approaches, while preserving the explainability of rules.

%\todo{Mihai -- I couldn't find a good place to discuss Goldberg's paper here without it feeling tacked on.  I cite it above...} % ms: Ok
% Industry has focused on rules because they are interpretable and maintainable (\todo{cite}), while academia has focused on neural approaches because they are able to generalize \todo{something else}.  Here, we incorporate both approach to benefit from the advantages of each.

% ms: removed. this is discussed in experiments
%\cite{Sabo2021RevisitingFR} applies the Few-Shot Learning paradigm, as formalized by \cite{Vinyals2016MatchingNF}, to the relation classification task in an attempt to better simulate the real-world use case. The authors propose a procedure to transform any supervised dataset into a few-shot dataset and apply it on the TACRED dataset \cite{zhang2017tacred}. On this new dataset, the top performing models on datasets such as \cite{Han2018FewRelAL, Gao2019FewRel2T} obtain a drastically lower performance, with typical scores on the 5-way 1-shot scenario averaging around 10\% F1. We use the same dataset in our empirical experiments, and show that our approach outperform state-of-the-art neural methods in the 1-shot scenario.

% It proposes certain modifications to the TACRED dataset and proceeds to show that models which achieved superhuman performance on datasets such as FewRel have a drop in performance of over 40\% F1 points  }

%% file: tex/problem.tex
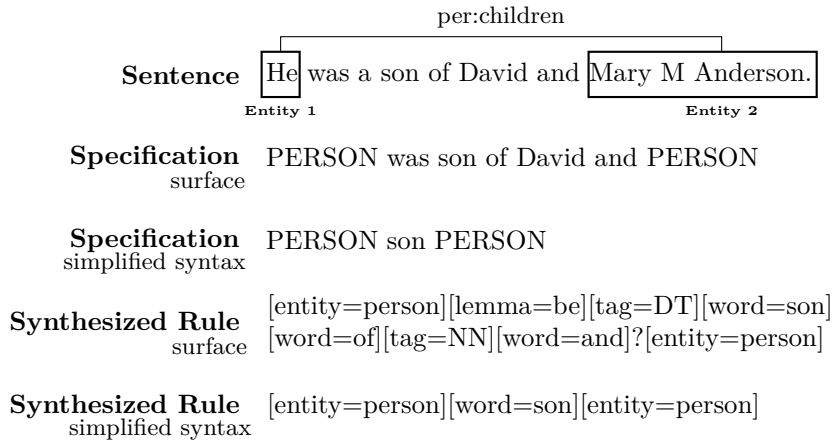
\begin{figure*}[t]
    \centering
%    \resizebox{\columnwidth}{!}{
      \input{figures/rule-example.tex}
%    }
    \caption{
        Example of input/output for our rule synthesis method.
        Our approach learns to generate individual rules that
        match a set of examples (or a {\em specification}),
        where each example is a highlighted span of text in a sentence.
        The ``simplified syntax'' lists the tokens contained on the shortest dependency path that connects the two entities.
        Note that this solution is not unique. 
    }\label{fig:sentence}
%    \vspace{-3mm}
  \end{figure*}

\section{Problem Statement}
In this effort, we address the problem of generating (or synthesizing) rules for IE from a few examples, 
satisfying two key constraints: 
we cannot assume experience with linguistics or machine learning from the users, 
and the approach must be domain-agnostic.
% Further, we require that the methodology be domain agnostic, and as close as possible to language agnostic (though for this first effort we focus on English only).
% Specifically we propose an approach for automatic rule generation using enumerative search, guided by a transformer-based language model for rules, and optimized using search-space pruning heuristics.

In our approach, the user is able to specify \textit{what} they need extracted by highlighting portions of text in sentences they have selected.
% In this paper we explore the problem of automatically generating rules based on human-selected sentences and human-defined specifications. 
For example, if a user is interested in extracting \textit{parent-child} relations, they might select a sentence such as \textit{\sentenceexample}, and from that sentence they might highlight \highlightedexample~as the content of interest (see Figure~\ref{fig:sentence}).\footnote{Here we are focusing on binary relation extraction, so while \textit{David} is also correct for this relation, it would be a separate specification sentence.}
Note that at no time in the process do they need to concern themselves with the underlying syntactic structure or language model.
This information forms the input to our method, which then searches for a rule that matches \textit{only} the highlighted part of the input sentence(s), such as the one shown in the figure.

Before we describe the actual algorithm, we introduce necessary terminology:

{\flushleft {\bf Specification:}} We call a (sentence, selection) pair, such as the one shown in Figure~\ref{fig:sentence}, a {\em specification}. 
We overload the term specification to refer to one or multiple such  pairs.
Note that the selection may be empty, as in the case of counter-examples, in which case a generated rule should match nothing. 
%\edit{For example, we can add the sentence \texttt{He is the son of John}, but \textit{without} any highlighted span. Then, the rule generated by the system will not match anything in this sentence. As such, instead of \texttt{[lemma=be]} in the surface synthesized rule, the system will choose \texttt{[word=was]}.}
% Note that 
%The input to the rule synthesis algorithm may be one or more such specifications.
The algorithm is required to generate a single rule that exactly matches  all highlights in the provided specification, and nothing more. 
 
% We call this ([sentence], [selection]) pair a specification. This specification (or specifications, if additional sentences/selections are provided) forms the input for the system, which then searches for a rule that, when applied to the input sentence(s), only matches the selection(s) and nothing more. For this example, such a rule could be "[word=is] [tag=DT] [tag=NN]".

{\flushleft {\bf Rules:}} We use Odinson \cite{valenzuela2020odinson}, 
a % highly optimized % ms: marketing
rule-based information extraction system, for rapidly evaluating a potential rule against the input. 
We chose Odinson because it brings two key advantages. First, Odinson rules are expressive; for example, a single rule can combine surface information with syntactic dependency paths. Second, the Odinson runtime engine is fast: its authors report that, due to its careful indexing of syntactic information, Odinson traverses syntax-based graphs six orders of magnitude faster than rule-based  systems that operate without such an index.

% This is one of the mechanism that helps guiding our search. 
As the search is informed by the syntax of Odinson, 
an important benefit of our approach is that every rule provided by the system is always a valid Odinson query. 
    Though Odinson is able to support rules which combine surface and syntax, for this initial effort we focus only on rules that rely on token sequences, which usually come from surface information.  We do, however, explore a linearization of syntactic paths, by using a sequence of tokens that consists of the tokens visited during a traversal of the shortest dependency path connecting the entities in the specification.
    For example, the shortest syntactic dependency path that connects {\em He} and {\em Mary M Anderson} in the sentence from Figure \ref{fig:sentence} contains the (unlabeled) dependencies: {\em son} $\rightarrow$ {\em He} and {\em son} $\rightarrow$ {\em Anderson},\footnote{We used CoreNLP dependencies \cite{manning2014stanford}.} which is linearized to preserve sentence order as: {\em He} {\em son} {\em Anderson}.
    This representation is referred to as \textit{simplified syntax} in Figure \ref{fig:sentence}.
    %Our proposed method outputs rules which operate over sequences of tokens. We explore two ways of generate sequences of tokens: (1) %surface sequences; or (2) sequences that come from the dependency parse. For example, 

% \footnote{Odinson is able to support both surface and syntax, and in fact to switch flexibly between representations within a rule.  We acknowledge that the ability to handle traversals of syntactic dependencies is critical, and plan to address this in future work.} 

{\flushleft {\bf Placeholder and state:}} 
Each search begins with a \textit{placeholder} (represented here as \placeholder), 
% and sometimes informally referred to as a \textit{hole}), % ms: let's not call it a hole...
which can be replaced with any valid rule element during the rule synthesis process.
We use the term {\em state} to refer to the information available at a given step of the rule generation; this information includes the current, intermediate form of the rule, as well as which parts of the specifications are matched. 
% The search begins with an empty rule containing a single placeholder (represented here as \placeholder).  
During rule generation, placeholders are iteratively expanded until we either (a) find a state that is a valid Odinson rule that satisfies the specification constraints, or (b) we reach a maximum number of steps, in which case no rule is produced. 
At each expansion, the algorithm determines the potential next states from the DSL, 
scores them based on their likelihood to be part of the completed rule, 
and adds them to a {\em priority queue} that is sorted in descending order of scores.  
The next state is then selected according to the queue. 
An example of possible expansion rules for \placeholder~ is 
given here:%\footnote{Please see Appendix \ref{sec:appendix_transitions} for a complete set of valid transitions.}

{\footnotesize
\begin{align*}
    \placeholder &\rightarrow \placeholder~\placeholder &&\qquad \text{(concatenation)}  \\
    \placeholder &\rightarrow [\placeholder] &&\qquad \text{(token constraint)}  \\
    \placeholder &\rightarrow \placeholder | \placeholder &&\qquad \text{(alternation)}  \\
    \placeholder &\rightarrow \placeholder\{?,*,+\} &&\qquad \text{(quantification)}
    % [\placeholder] &\rightarrow [!\placeholder] &&\qquad \text{(not constraint)} \\
    % [\placeholder] &\rightarrow [\placeholder ~ \& ~ \placeholder] &&\qquad \text{(and constraint)} \\
    % [\placeholder] &\rightarrow [\placeholder=\placeholder] &&\qquad \text{(token constraint)} \\
    % &&\dots
    % [\placeholder] &\rightarrow [\placeholder=\placeholder] &&\qquad \text{(token constraint)} \\
    % [\placeholder] &\rightarrow [!\placeholder] &&\qquad \text{(not constraint)} \\
    % [\placeholder] &\rightarrow [(\placeholder \& \placeholder)] &&\qquad \text{(and constraint)} \\
    % [\placeholder] &\rightarrow [(\placeholder | \placeholder)] &&\qquad \text{(or constraint)} \\
\end{align*}
}
\vspace{-3mm}

Note that the search space grows exponentially with depth, making a brute-force approach intractable.
For example, attempting brute force to generate \texttt{[entity=person] [lemma=be] [tag=DT] [word=son] [word=of] [tag=NN]? [word=and]? [entity=person]}, 
one of the solutions to our previously introduced (sentence, selection) input pair, 
yields $\gg$1 billion states to explore. 
% Even for this small example the search space gets prohibitively large. As such, we need an efficient way of exploring the search space.
Therefore, the efforts to both \textit{prioritize} which states to explore first (Section \ref{sec:scorer}) and 
to \textit{prune} portions of the search tree that cannot yield a correct rule (Section \ref{sss:prune_search_space}) are crucial.

%% file: figures/rule-example.tex
\begin{tikzpicture}
  [
  lf/.style={
    anchor=east
  },
  lr/.style={
    anchor=west
  },
]

\newcommand\rc{0.1}
\newcommand\hc{1.1}
% 1
\node[lf] at (0,10 -\hc*0) {\textbf{Sentence}};

\node[lr] (s) at (\rc,10 -\hc*0) {He was a son of David and Mary M Anderson.};
\draw[thick] ($(s) - (3.15,-.3)$) rectangle ($(s) - (3.65,  .3)$);
\node (e1) at ($(s) - (3.4,0.5)$) {\tiny \textbf{Entity 1}};

\draw[thick] ($(s) - (-3.65,-.3)$) rectangle ($(s) - (-0.65,  .3)$);
\node (e2) at ($(s) - (-2.4,0.5)$) {\tiny \textbf{Entity 2}};

\draw ($(e1) + (0,0.8)$) -- ($(e1) + (0,1) $)
-- node[above] {\small per:children} ++(5.8,0)
-- ($(e2)+ (0,1)$) -- ($(e2) + (0,0.8)$) ;

\node[lf] (sp1) at (0,10-\hc*1) {\textbf{Specification}};
\node at ($ (sp1) - (-0.7, 0.3) $) {\small surface};

\node[lr] at (\rc,10-\hc*1) {PERSON was son of David and PERSON};

\node[lf] (sp2) at (0,10-\hc*2) {\textbf{Specification}};
\node at ($ (sp2) - (0.0, 0.3) $) {\small simplified syntax};

\node[lr] at (\rc,10-\hc*2) {PERSON son PERSON};

\node[lf] (sr1) at (0,10-\hc*3) {\textbf{Synthesized Rule}};
\node at ($ (sr1) - (-1.12, 0.3) $) {\small surface};

\node[lr, xshift=-0.2cm] at (\rc,10-\hc*3) {
  \begin{tabular}{l}
    {[}entity=person{]}{[}lemma=be{]}{[}tag=DT{]}{[}word=son{]} \\
    {[}word=of{]}{[}tag=NN{]}{[}word=and{]}?{[}entity=person{]} \\
  \end{tabular}
};

\node[lf] (sr2)  at (0,10-\hc*4) {\textbf{Synthesized Rule}};
\node at ($ (sr2) - (-0.47, 0.3) $) {\small simplified syntax};

\node[lr, xshift=-0.2cm] at (\rc,10-\hc*4) {
  \begin{tabular}{l}
    {[}entity=person{]}{[}word=son{]}{[}entity=person{]}\\
  \end{tabular}
};

\end{tikzpicture}

%% file: tex/method.tex
% \begin{figure*}[ht]
%     \centering
%     \includegraphics[width=0.75\textwidth]{figures/odinsynth_architecture9.png}
%     \vspace{-3mm}
%     \caption{\footnotesize{
%         Conceptually, the algorithm consists of three main steps: \textbf{expand} the current state, \textbf{score} the possible next states, and \textbf{select} the state with the highest score. In the beginning, we start from the placeholder ($\square$) and expand it (1). We score each expansion (2) and transition to the one with the highest score (3). We then expand and score again (4,5). At this point, notice that the state with the highest score is $\square~\square$, which is in a different subtree than the current state ($\square?$). The enumerative searcher, with Branch \& Bound as the underlying algorithm, transitions to the state with the highest score, regardless of the current state position (6). This process repeats (7-9) until we arrive at a state that is a valid rule (last block), or we reach a maximum number of steps and stop.
%         % \todo{Switch to simpler figure. Also, step 2 has a spurious arrow under the one with 0.21.}
%     }}
%     \label{fig:architecture}
%     \vspace{-3mm}
% \end{figure*}

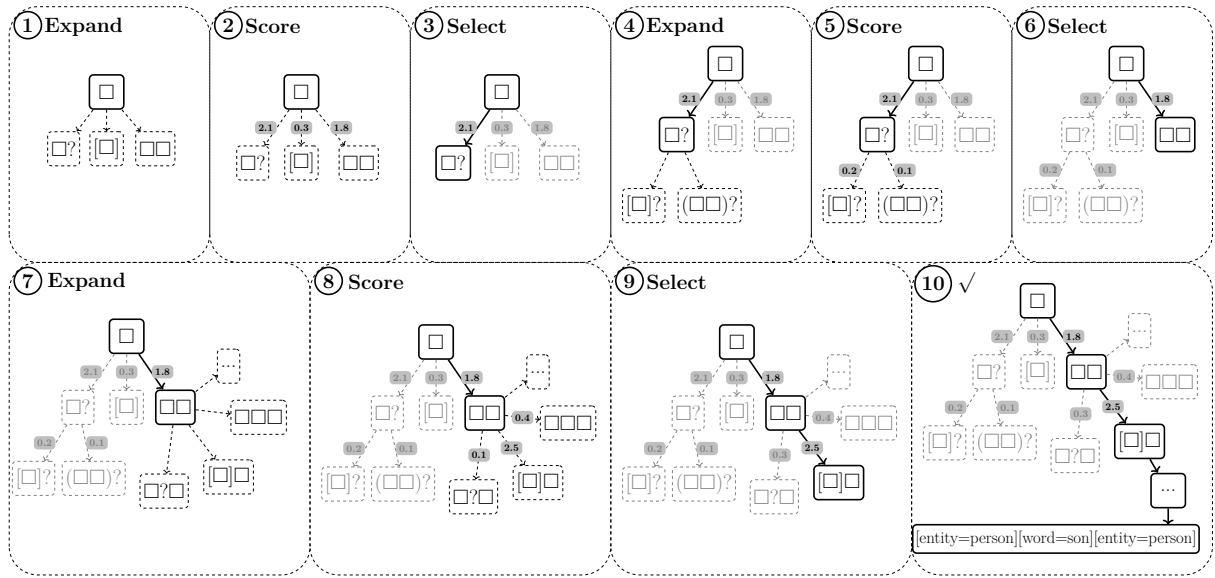
\begin{figure*}[ht]
    \centering
    \resizebox{!}{0.98\columnwidth}{
    \input{figures/algorithm-example.tex}
    }
    \caption{
        Conceptually, the algorithm consists of three main steps: \textbf{expand} the current state, \textbf{score} the possible next states, and \textbf{select} the state with the highest score. In the beginning, we start from the placeholder ($\square$) and expand it (1). We score each expansion (2) and transition to the one with the highest score (3). We then expand and score again (4,5). At this point, notice that the state with the highest score is $\square~\square$, which is in a different subtree than the current state ($\square?$). The enumerative searcher, with Branch \& Bound as the underlying algorithm, transitions to the state with the highest score, regardless of the current position (6). This process repeats (7-9) until we arrive at a state that is a valid rule (10), or we reach a maximum number of steps and stop. 
    }
    \label{fig:architecture}
\end{figure*}

\section{Method}
% Our approach consists of a search algorithm, branch and bound, with 
% In a nutshell, our proposed approach for rule generation uses transformer-guided enumerative search, further optimized with search-space pruning heuristics.
In a nutshell, our proposed approach for rule generation uses enumerative search that is guided by a transformer-based scoring mechanism, and is optimized using search-space pruning heuristics.
%We propose an approach for automatic rule generation using enumerative search, guided by a transformer-based language model for rules, and optimized using search-space pruning heuristics.
Our transformer model scores each potential next state, given the current state, such that the number of states to be explored is minimized. 
Specifically, our system consists of two main components:

{\flushleft A \textbf{searcher}} (Section \ref{sec:searcher}), with Branch and Bound~\cite{bnb} as the underlying algorithm. The searcher uses the scores assigned by the scorer (below) to determine the order of exploration, choosing the state with the highest score, 
\textit{regardless of its position in the search tree}. 
As such, it is important for the scorer to assign high scores to states that are in the sub-tree 
that leads to the desired final rule, and lower scores to all other states;

{\flushleft A \textbf{scorer}} (Section \ref{sec:scorer}), with a transformer backbone that is initialized with a pretrained model, but fine-tuned through self-supervision, i.e., over {\em automatically generated rules}.
The scorer uses the current state and the specification to score each potential next state.

\subsection{Enumerative Searcher}
\label{sec:searcher}
The searcher is responsible for exploring the states in priority order (as determined by the scorer), and deciding if a given state is successful (i.e., it is a valid query and correctly extracts the requested highlighted words and nothing more).
The search space can be interpreted as a tree, where the root is the initial candidate solution and the children of a node $n$ are the candidate solutions that the node $n$ could expand to.
Given this, the searcher can be seen as iteratively applying a sequence of three operations: (a) \textbf{Expand} the current state according to the DSL grammar,\footnote{We always expand the leftmost placeholder first.} 
(b) \textbf{Score} each expanded candidate next state and insert them into the priority queue, and (c) \textbf{Select} from the queue the state with the highest score to be the next state. We repeat this process until we find a solution or we reach our step limit.

Figure \ref{fig:architecture} shows a detailed example for the input sentence \textit{\sentenceexample}, with \textit{\highlightedexample} as the highlighted span or selection.   
In this example, we generate one possible solution: \texttt{[entity=person] [word=son] [entity=person]}, which comes from the linearization of the dependency path between the two {\tt person} named entities.

\subsubsection{Pruning the search space}
\label{sss:prune_search_space}
While the scorer determines the order of exploration, this is complemented by techniques for greatly pruning the search space to be considered. 
% to reduce the number of nodes that need to be considered. 
In particular, adapting the techniques of \cite{lee2016synthesizing} to our use case, 
we prune states for which the \textit{least restrictive} rule that could result from this state cannot completely match the highlighted specification, as nothing created from that subtree can be a solution. 
Consider the state: \texttt{[entity=person] [word=born] $\placeholder$}.
The least restrictive rule resulting from this state would be one which matches the word \textit{person}, 
followed by \textit{born} and then 0 or more (unrestricted) tokens. 
If such a rule cannot completely match the highlighted tokens, then a valid solution cannot be found in that subtree so we prune the branch.
% \todo{BUT this rule does match the correct solution, no? Is this the right example here?}
% because that tree cannot contain a solution.
% e utilize the following strategies for pruning the search space prior to scoring:

% \paragraph{Under-approximation} Given a current state, consider the most restrictive or conservative form of a rule that could possibly result from expanding this state. If this under-approximation matches something outside the specification, i.e., parts that were not highlighted, then we should not explore that branch further, so the branch is pruned. 

% \paragraph{Over-approximation} given a current state, If the least restrictive rule that could result from this state cannot completely match the highlighted specification, then that tree cannot contain a solution. 
% \footnote{An example of the least restrictive rule for "[word=is] [tag=$\placeholder$]" is a rule that matches everything after a word is.}

% \footnote{It cannot contain a solution because a solution would completely match the highlighted parts, but if we could not match them with the over approximation query, then no expansion can.} 
% \todo{add more!  what exactly does "most" and "least" restrictive mean here? specifically what do we do?}

% \todo{Robert: an example would be nice here, if space permits}
% \vspace{-1mm}
\subsection{Scorer}
\label{sec:scorer}

The Scorer assigns a numerical value to states to establish the order of exploration.
% Intuitively, it should give the highest score to the state that is the correct continuation of the previous state. 
We explore two variants: as a baseline, we implement a \textbf{static} variant based on the components of a given state, and a \textbf{contextual} variant based on a self-supervised model that takes the current context into account.%\footnote{Here we describe our language model as self-supervised.  There was no annotation involved (i.e., it is not supervised), instead we train what is essentially a masked language model over a large, automatically generated corpus of random rules.} 

\subsubsection{Static weights}
\label{sec:static}
For this baseline, the score of a state is solely determined by its components. The cost of each state is constructed by summing the cost of its components with the cost of its node. For example, the cost of \placeholder ~ \placeholder ~ (concatenation) is: 
% calculated as the sum of the costs of each of its components plus the cost associated with the operation itself: 
% \mbox{\small $score(\square ~ \square) = -1 * (cost(\square) + cost(\square) + cost(concatenation))$}
\mbox{\small $cost(\square ~ \square) = cost(\square) + cost(\square) + cost(concatenation)$}

% \begin{small}
%     \begin{align*}
%     score(\square ~ \square) &= -cost(\square ~ \square) \\
%     cost(\square ~ \square) &= cost(\square) + cost(\square) + cost(concatenation)
%     \end{align*}
% \end{small}

The costs for each operation were hand-tuned based on intuition (e.g., exploring negation takes a very long time as you need to consider everything some constrain \textit{cannot} be, thus negation is given a higher cost), then optimized on a small  external development set of sentences and specifications.
% Notice that there is nothing dynamic in the above formula. 

In addition to the hand-tuned nature of the static scorer, there are two main limitations.
First, %the static weights mean that 
a given state will always receive the same score regardless of the sentence context or the previous state. 
Second, states with more components in their underlying pattern inherently have a higher cost because the score is
summative. 
% derived by adding the cost of the components to the cost of the operation.  
%Thus, \texttt{[tag=JJ] [word=great] [word=dragons]} will always have a smaller score than \texttt{[tag=JJ] [word=green]}, regardless of context.
% \footnote{This is because the score of a state is constructed as its negative cost, where the cost is derived by adding the cost of each of its components to the cost of the node.} 
This is undesirable, as the 
states which expand to a solution should score higher than those which cannot, regardless of length.
% score of a state which can expand to a solution should be higher than that of a state which can not, regardless of length.
% A second limitation is that the score of a state is constructed as its negative cost, where the cost is derived by summing the cost of each of its components. This implies that longer states, such as 

% \begin{figure}[ht]
%     \centering
%     \includegraphics[width=1\columnwidth]{figures/static_scorer_example2.png}
%     \caption{
%         Illustration of limitations of the static scorer. Consider two specifications: \textit{s1} and \textit{s2}, the current state \textit{cs}, and two possible next states: \textit{nps1} and \textit{nps2}. The static scorer will assign the same value to both \textit{nps1} and \textit{nps2}, regardless of the specification, as described in Section \ref{sec:static}.
%         % \todo{sentence or specification here?}
%         % , even though a transition from \textit{cs} to \textit{nps1} is good for \textit{s1}, but not for \textit{s2}. 
%     }
%     \label{fig:static_scorer_limitations}
% \end{figure}

% Both limitations are addressed next with the dynamic weights approach. Such an approach uses the sentence on which the final pattern will be applied to guide the score. As such, it is not constrained to give lower scores to states with long underlying pattern.
% This limitations is addressed next with the dynamic weights approach.

\subsubsection{Score augmentation based on estimation}
\label{sec:reward}
%\todo{Maybe trim this?} % ms: no, this is good... Keep if we can
To supplement the score from the static weights, we introduce an additional 
score that estimates how well a current (incomplete) rule matches the specification so far.
% partial reward.
% While in an incomplete rule (i.e., one that still contains one or more placeholders),
% cannot be a solution, 
% we can determine a preliminary estimate of how well it matches the specification.  
For this, we remove all components of an incomplete rule that contain a placeholder 
and apply the remainder of the rule to the specification. We then boost the state's score
for each specification token that is correctly matched, 
and penalize for each token incorrectly matched.

% ms: redundant with the text above
%Specifically, 
%The reward consists in the number of tokens 
%matched by an incomplete rule.
%By incomplete we mean a rule that contains one or more
%\placeholder.
%In order to be able to match 
%the rule first we need to remove the holes.

For example, for a rule such as \texttt{[entity=person] [tag=NN] [word=\placeholder]},
we remove incomplete components, resulting in the rule: 
\texttt{[entity=person] [tag=NN]}, which is matched against the specification.
For each highlighted token matched, 
% =======
% for a rule such as \texttt{[word=\placeholder] [word=son] [word=person]},
% we remove incomplete components, resulting in the rule: 
% \texttt{[word=son] [word=person]}.
% We then match the new rule against the specification, 
% and for each highlighted token matched, 
% >>>>>>> 5b0d766de3c0bbde4d6b4dd385f95477492c64b7
the function adds $1$ to the score.
For matches outside the highlight, the function adds $-1$.
% This score is returned and used to augment the score given by the static scorer.
% The partial reward function then returns the sum of the values 
% obtained by matching all examples in the stash.
% \footnote{\todo{if we add this to the contextual weights, say so (did not help)}}
We observe that using this score augmentation favors more concrete constraints,\footnote{More word constraints instead of tag or lemma constraints} which help ground the rule to more 
lexical artifacts, but may hinder generalization. 
% ms: move sentence below to results.
% We observed empirically that the partial reward did not improve generalization, but it increased the number of rules that were found.

\subsubsection{Contextual weights}
To address the limitations of the static weights, 
we propose a \textit{contextual} scorer that utilizes the current 
context (i.e., the specification and the current state), 
to determine the cost of a candidate next state.
Unlike our score augmentation, here we use the 
full specification, not just what is matched at a given time.
% is also determined by the sentence and by the previous state. 
% This makes possible for the system to  <..>

For this scorer, we use a transformer-based encoder to score 
each (current state, next potential state, specification) input.
Intuitively, this score is the likelihood that the next potential state is better than the current state, 
which allows the scores to be comparable across all levels in the search tree. 
% input into a single vector, and a linear layer to map the vector to a single numerical value. 
% A state which eventually expands to a solution should have a higher score than states that do not. 
%to obtain a single numerical value. 
%This numerical value represents the score. Intuitively, the state with the highest score should be more likely to be on the correct subtree.

% \begin{itemize}
%     \item Static approach, where a pattern gets the same score, regardless of the sentence. The cost of each rule is constructed by summing the cost of its components. For example, the transition from \placeholder to \placeholder \placeholder (concatenation) has a cost of $cost(\placeholder) + cost(\placeholder) + cost(concatenation)$.
%     \item Dynamic approach, where the cost of a pattern is also determined by the sentence. We used a transformer-based encoder to obtain a single vector from the input and a linear layer to obtain a single numerical value.
% \end{itemize}

% The architecture is shown in Figure \ref{fig:scorer}. 
Our contextualized scorer consists of a variant of BERT~\cite{DBLP:journals/corr/abs-1810-04805,turc2019wellread}\footnote{We experiment with multiple pre-trained variants of BERT, introduced in \cite{turc2019wellread}} with a linear layer on top. 
The BERT encoder input is a concatenation of: 
(1) linearized AST of the current state (e.g., \placeholder), (2) linearized AST of the next potential state (\placeholder?), 
(3) and the (sentence, selection) specification.
% , which is a sentence with a highlighted portion (She \textit{is an American} actress and singer). 
Since these concatenated components are fundamentally different, 
we differentiate between them by using different token type ids in the encoder. 
The tokens from the current state have a token type id of $1$, and tokens of the next potential state have $2$. 
We further differentiate between the highlighted and non-highlighted portions of the specification 
text in the same way, with token type ids $3$ and $4$, respectively.

\subsubsection{Multiple sentences}
So far we have used only a single sentence in our specification examples. 
Nevertheless, our system can handle multiple sentences and their highlights. 
We require the enumerative searcher to find a rule that would satisfy \textit{all} the constraints for all sentences in the specification.
When scoring, we score a (current state, next potential state, single-sentence specification) triple, 
and then average over all sentences in the specification to obtain a final score for the (current state, next potential state) transition.
%\footnote{We briefly experimented with adding an LSTM between the BERT encoder and the scorer to aggregate the encodings of each (current state, next potential state, sentence specification) triple into one vector. However, we didn't observe a noticeable improvement, so we omitted it for simplicity.}

\subsubsection{Training}
\label{sss:training}

Unlike the static scorer, the neural guiding function of the contextual scorer needs to be trained, 
which we do with self-supervision.  
Because there is no large corpus of Odinson rules,
we artificially generate one with random spans of text that we randomly manipulate into rules.
% by randomly selecting spans of text in a large, unlabeled corpus, and then randomly 
% modify the \textit{representation} of the tokens in the span to create a random rule that matches the span.
Our random-length text spans are chosen from the UMBC corpus \citelanguageresource{han2013umbc}.
Each token in this span is then randomly manipulated into an Odinson token constraint based on either word, lemma, or part-of-speech.
For example, a span such as \textit{the dog barked} might be converted to \texttt{[tag=DT] [word=dog] [lemma=bark]}.
Then, to expose the model to additional rule components (e.g., alternation, quantifiers), we add further manipulations, again with randomization.
To add alternations, we build a temporary query by replacing one of the token constraints with a wildcard that can match \textit{any} token and
query the corpus for an additional sentence that has different content in that position.  This new content is added as an alternation.
For example, with the temporary version of the above query \texttt{[tag=DT] [word=dog] []},\footnote{Note 
that the Odinson wildcard, \texttt{[]} looks similar to, but is not the same as our placeholder, \placeholder.} we might find \textit{A dog runs}, resulting 
in the following alternation: \texttt{[tag=DT] [word=dog] ([lemma=bark]|[lemma=run])}. %\todo{Becky: I am not sure "lemma/bark" is valid in Odinson...}
To add a quantifier (i.e., \texttt{*}, \texttt{+}, or \texttt{?}), we select a token to modify and a quantifier to add, 
and check the corpus to ensure that the addition of the quantifier yields additional results. 

After generating each random rule, we build a corresponding specification by 
querying the UMBC corpus: the retrieved sentences and their matched spans constitute %the (sentence, selection) pairs of the 
specification.
However, having a specification and the corresponding rule is not enough to train our model.  We also need a correct sequence of 
transitions from the initial placeholder to the final rule.
For this, we use an Oracle to generate the shortest sequence of transitions, 
which we consider to be the correct sequence for our purposes. 
This sequence of transitions, together with the specification, forms the training data for our model.
Note that we train \textit{only} on this data, i.e., after this self-supervised training process the transformer's weights are fixed.
% Given this data, there are different ways to frame the rule-synthesis task, such as binary .
% random selection of sentences, random weights, run system
% This means that for each  we have $cost(\placeholder)=1$ and $cost(concatenation) = 3$, yielding $cost(\placeholder \placeholder) = 5$, while for other runs we can have $cost(\placeholder)=2$ and $cost(concatenation) = 6$, yielding $cost(\placeholder \placeholder) = 10$
% We view the training data as a binary prediction task. 
% Given a specification set, the current state, and every valid candidate next state, we want to predict 1 for each (current state, correct next state, specification) triple and 0 for all the others. 
We train using the cross-entropy loss and with a cyclical learning rate, as suggested by \cite{smith2017cyclical}. 
Further, we employ a curriculum learning approach \cite{bengio2009curriculum,stretcu2019curriculum}, 
splitting the training data by sentence length and by pattern length. 
We did not tune our hyperparameters because we want to maintain the synthesis approach domain agnostic, rather than fine-tuning on a specific task. Our results indicate that this is the case.
% (listed, along with additional training procedure details, in Appendix \ref{sec:hyperparameters}). 

% in this scenario

% \paragraph{Ranking}
% A second way of viewing the training data is that of a ranking task. 
% Using a max-margin loss, we enforce that the score of the correct next state should be higher than the score of any other candidate next state.  
% Here we use the max-margin loss. %in this scenario.

% \paragraph{Hybrid training}
% Finally, we combine both of these training approaches into a hybrid approach, where our system is initially trained with the binary prediction task, then we switch the objective to ranking. \todo{Why?  defend the idea with some intuition} \todo{Updates. Curriculum learning. Smaller BERT vs larger BERT?}

%% file: figures/algorithm-example.tex
\begin{tikzpicture}
  [cel/.style={
    draw,
    dashed,
    rounded corners,
    minimum height=1.2cm,
    thick,
    color=gray,
  },
  cel_enabled/.style={
    draw,
    rounded corners,
    minimum height=1.2cm,
    minimum width=1.2cm,
    ultra thick
  },
  score/.style={
    % draw,
    rounded corners,
    fill=lightgray
  },
  disabled_edge/.style={
    color=gray,
    dashed,
    thick,
    ->
  },
  enabled_edge/.style={
    ultra thick,
    ->
  },
  box/.style={
    dashed,
    rounded corners=1cm,
    thick,
  },
  numbers/.style={
    draw,
    circle,
    ultra thick,
    yshift=-.15cm,
    xshift=.15cm
  }
]

% 1
\node [cel_enabled, yshift=-2cm, xshift=0.4cm] at(0,14) {\huge $\square$} [level distance=2cm]
child {node [cel, color=black] {\huge $\square$?}
  edge from parent [disabled_edge, color=black]}
child {node [cel, color=black] {\huge [$\square$]}
  edge from parent [disabled_edge, color=black]}
child {node [cel,xshift = 0.2cm, color=black] {\huge $\square$$\square$}
  edge from parent [disabled_edge, color=black]};

% 2
\node [cel_enabled, yshift=-2cm, xshift=0.2cm] at(7,14) {\huge $\square$} [level distance=2.5cm]
child {
  node [cel, xshift=-0.2cm, color=black] {\huge $\square$?}
  edge from parent [disabled_edge, color=black]
  node [score, left] {\textbf{2.1}}}
child {
  node [cel, xshift=0cm, color=black] {\huge [$\square$]}
  edge from parent [disabled_edge, color=black]
  node [score] {\textbf{0.3}}}
child {
  node [cel, xshift=0.5cm, color=black] {\huge $\square$$\square$}
  edge from parent [disabled_edge, color=black]
  node [score,right] {\textbf{1.8}}};

% 3
\node [cel_enabled, yshift=-2cm, xshift=0.2cm] at(14,14) {\huge $\square$} [level distance=2.5cm]
child {node [cel_enabled, xshift=-0.2cm] {\huge $\square$?}
  edge from parent [enabled_edge] node [score, left] {\textbf{2.1}}}
child {node [cel, xshift=0cm] {\huge [$\square$]}
  edge from parent [disabled_edge] node [score] {\textbf{0.3}}}
child {node [cel, xshift=0.5cm] {\huge $\square$$\square$}
  edge from parent [disabled_edge] node [score,right] {\textbf{1.8}}};

% 4
\node [cel_enabled, yshift=-1cm, xshift=1cm] at(21,14) {\huge $\square$} [level distance=2.5cm]
child {
  node [cel_enabled, xshift=-0.2cm] {\huge $\square$?}
  edge from parent [enabled_edge]
  child {
    node [cel, xshift=-0.4cm, color=black] {\huge [$\square$]?}
    edge from parent [disabled_edge, color=black]
  }
  child {
    node [cel, xshift=0.4cm, color=black] {\huge ($\square$$\square$)?}
    edge from parent [disabled_edge, color=black]
  }
  node [score, left] {\textbf{2.1}}}
child {
  node [cel] {\huge [$\square$]}
  edge from parent [disabled_edge]
  node [score] {\textbf{0.3}}
}
child {
  node [cel, xshift=0.2cm,] {\huge $\square$$\square$}
  edge from parent [disabled_edge]
  node [score,right] {\textbf{1.8}}
};

% 5
\node [cel_enabled, xshift=1cm, yshift=-1cm] at(28,14) {\huge $\square$} [level distance=2.5cm]
child {
  node [cel_enabled, xshift=-0.2cm] {\huge $\square$?}
  edge from parent [enabled_edge]
  child {
    node [cel, xshift=-0.4cm, color=black] {\huge [$\square$]?}
    edge from parent [disabled_edge, color=black]
    node[score, left] {\textbf{0.2}}
  }
  child {
    node [cel, xshift=0.4cm, color=black] {\huge ($\square$$\square$)?}
    edge from parent [disabled_edge, color=black]
    node[score,right] {\textbf{0.1}}
  }
  node [score, left] {\textbf{2.1}}}
child {
  node [cel] {\huge [$\square$]}
  edge from parent [disabled_edge]
  node [score] {\textbf{0.3}}
}
child {
  node [cel, xshift=0.2cm,] {\huge $\square$$\square$}
  edge from parent [disabled_edge]
  node [score,right] {\textbf{1.8}}
};

% 6
\node [cel_enabled, xshift=1cm, yshift=-1cm] at(35,14) {\huge $\square$} [level distance=2.5cm]
child {
  node [cel, xshift=-0.2cm] {\huge $\square$?}
  edge from parent [disabled_edge]
  child {
    node [cel, xshift=-0.4cm] {\huge [$\square$]?}
    edge from parent [disabled_edge]
    node[score, left] {\textbf{0.2}}
  }
  child {
    node [cel, xshift=0.4cm] {\huge ($\square$$\square$)?}
    edge from parent [disabled_edge]
    node[score,right] {\textbf{0.1}}}
  node [score, left] {\textbf{2.1}}
}
child {
  node [cel] {\huge [$\square$]}
  edge from parent [disabled_edge]
  node [score] {\textbf{0.3}}}
child {
  node [cel_enabled, xshift=0.2cm] {\huge $\square$$\square$}
  edge from parent [enabled_edge] [grow=330, level distance=2cm]
  node[score,right] {\textbf{1.8}}
};

% 7

 \node [cel_enabled, xshift=0.1cm, yshift=-1.6cm] at(1,5) {\huge $\square$} [level distance=2.5cm]

child {
  node [cel, xshift=-0.2cm] {\huge $\square$?}
  edge from parent [disabled_edge]
  child {
    node [cel, xshift=-0.8cm] {\huge [$\square$]?}
    edge from parent [disabled_edge]
    node[score, left] {\textbf{0.2}}}
  child {
    node [cel, xshift=-0.2cm] {\huge ($\square$$\square$)?}
    edge from parent [disabled_edge]
    node[score,right] {\textbf{0.1}}}
  node [score, left] {\textbf{2.1}}}
child {
  node [cel] {\huge [$\square$]}
  edge from parent [disabled_edge]
  node [score] {\textbf{0.3}}}
child {
  node [cel_enabled, xshift=0.2cm, color=black] {\huge $\square$$\square$} [grow=330, level distance=2cm]
  edge from parent [enabled_edge]
  child {
    node [cel, xshift=-1cm, color=black] {\huge $\square$?$\square$}
    edge from parent [disabled_edge, color=black]
  }
  child {
    node [cel, yshift=-0.8cm, xshift=0.5cm, color=black] {\huge [$\square$]$\square$}
    edge from parent [disabled_edge, color=black]
  }
  child {
    node [cel, xshift=0.8cm, color=black] {\huge $\square$$\square$$\square$}
    edge from parent [disabled_edge, color=black]
  }
  child {
    node [cel, xshift=-1cm, yshift=0.5cm, color=black] {\huge ...}
    edge from parent [disabled_edge, color=black]}
  node [score,right] {\textbf{1.8}}
};

% 8
\node [cel_enabled, yshift=-1.8cm, xshift=0.4cm] at(11.5,5) {\huge $\square$} [level distance=2.5cm]
child {
  node [cel, xshift=-0.2cm] {\huge $\square$?}
  edge from parent [disabled_edge]
  child {
    node [cel, xshift=-0.8cm] {\huge [$\square$]?}
    edge from parent [disabled_edge]
    node[score, left] {\textbf{0.2}}}
  child {
    node [cel, xshift=-0.2cm] {\huge ($\square$$\square$)?}
    edge from parent [disabled_edge]
    node[score,right] {\textbf{0.1}}}
  node [score, left] {\textbf{2.1}}}
child {
  node [cel] {\huge [$\square$]}
  edge from parent [disabled_edge]
  node [score] {\textbf{0.3}}}
child {
  node [cel_enabled, xshift=0.2cm, color=black] {\huge $\square$$\square$} [grow=330, level distance=2cm]
  edge from parent [enabled_edge]
  child {
    node [cel, xshift=-1cm, color=black] {\huge $\square$?$\square$}
    edge from parent [disabled_edge, color=black]
    node [score] {\textbf{0.1}}}
  child {
    node [cel, yshift=-0.8cm, xshift=0.5cm, color=black] {\huge [$\square$]$\square$}
    edge from parent [disabled_edge, color=black]
    node [score] {\textbf{2.5}}}
  child {
    node [cel, xshift=0.8cm, color=black] {\huge $\square$$\square$$\square$}
    edge from parent [disabled_edge, color=black]
    node [score] {\textbf{0.4}}}
  child {
    node [cel, xshift=-1cm, yshift=0.5cm, color=black] {\huge ...}
    edge from parent [disabled_edge, color=black]}
  node [score,right] {\textbf{1.8}}
};

% 9
\node [cel_enabled, yshift=-1.8cm, xshift=0.4cm] at(22,5) {\huge $\square$} [level distance=2.5cm]
child {
  node [cel, xshift=-0.2cm] {\huge $\square$?}
  edge from parent [disabled_edge]
  child {
    node [cel, xshift=-0.8cm] {\huge [$\square$]?}
    edge from parent [disabled_edge]
    node[score, left] {\textbf{0.2}}}
  child {
    node [cel, xshift=-0.2cm] {\huge ($\square$$\square$)?}
    edge from parent [disabled_edge]
    node[score,right] {\textbf{0.1}}}
  node [score, left] {\textbf{2.1}}}
child {
  node [cel] {\huge [$\square$]}
  edge from parent [disabled_edge]
  node [score] {\textbf{0.3}}}
child {
  node [cel_enabled, xshift=0.2cm] {\huge $\square$$\square$} [grow=330, level distance=2cm]
  edge from parent [enabled_edge]
  child {
    node [cel, xshift=-1cm] {\huge $\square$?$\square$}
    edge from parent [disabled_edge]
    node [score] {\textbf{0.3}}}
  child {
    node [cel_enabled, yshift=-0.8cm, xshift=0.5cm] {\huge [$\square$]$\square$}
    edge from parent [enabled_edge]
    node [score] {\textbf{2.5}}}
  child {
    node [cel, xshift=0.8cm] {\huge $\square$$\square$$\square$}
    edge from parent [disabled_edge] node [score] {\textbf{0.4}}}
  child {
    node [cel, xshift=-1cm, yshift=0.5cm] {\huge ...}
    edge from parent [disabled_edge]}
  node [score,right] {\textbf{1.8}}
};

% 10
\node [cel_enabled, xshift=0.4cm, yshift=-0.35cm] at(32.5,5) {\huge $\square$} [level distance=2.5cm]
child {
  node [cel, xshift=-0.2cm] {\huge $\square$?}
  edge from parent [disabled_edge]
  child {
    node [cel, xshift=-0.8cm] {\huge [$\square$]?}
    edge from parent [disabled_edge]
    node[score, left] {\textbf{0.2}}}
  child {
    node [cel, xshift=-0.2cm] {\huge ($\square$$\square$)?}
    edge from parent [disabled_edge]
    node[score,right] {\textbf{0.1}}}
  node [score, left] {\textbf{2.1}}}
child {
  node [cel] {\huge [$\square$]}
  edge from parent [disabled_edge]
  node [score] {\textbf{0.3}}}
child {
  node [cel_enabled, xshift=0.2cm] {\huge $\square$$\square$} [grow=330, level distance=2cm]
  edge from parent [enabled_edge]
  child {
    node [cel, xshift=-1cm] {\huge $\square$?$\square$}
    edge from parent [disabled_edge]
    node [score] {\textbf{0.3}}}
  child {
    node [cel, yshift=-0.8cm, xshift=0.5cm, ultra thick, solid, color=black] {\huge [$\square$]$\square$} [grow=300, level distance=2cm]
    edge from parent [enabled_edge]
    child {
      node (secondtolast) [cel_enabled] {\huge ...} [grow=270, level distance=1.7cm] edge from parent [enabled_edge]
      child [anchor = east, child anchor=north east]  {
        node [draw,rounded corners, minimum height=1.25cm, minimum width=1.2cm, ultra thick, xshift=1.1cm, scale=0.8, fill=white] {\huge [entity=person][word=son][entity=person]}
        edge from parent [draw=none]}
    }
    node [score] {\textbf{2.5}}}
  child {
    node [cel, xshift=0.8cm] {\huge $\square$$\square$$\square$}
    edge from parent [disabled_edge]
    node [score] {\textbf{0.4}}}
  child {
    node [cel, xshift=-1cm, yshift=0.5cm] {\huge ...}
    edge from parent [disabled_edge]}
  node [score,right] {\textbf{1.8}}
};
\draw[ultra thick, ->] (secondtolast) -- ++(0,-1.2);

\draw[box] (-3,6) rectangle (4,15);
\draw[box] (4,6) rectangle (11,15);
\draw[box] (11,6) rectangle (18,15);
\draw[box] (18,6) rectangle (25,15);
\draw[box] (25,6) rectangle (32,15);
\draw[box] (32,6) rectangle (39,15);

\draw[box] (-3,-5) rectangle (7.5,6);
\draw[box] (7.5,-5) rectangle (18,6);
\draw[box] (18,-5) rectangle (28.5,6);
\draw[box] (28.5,-5) rectangle (39,6);

\node [numbers] at (-2.5, 14.5) {\huge \textbf{1}};
\node [numbers] at (4.5, 14.5) {\huge \textbf{2}};
\node [numbers] at (11.5, 14.5) {\huge \textbf{3}};
\node [numbers] at (18.5, 14.5) {\huge \textbf{4}};
\node [numbers] at (25.5, 14.5) {\huge \textbf{5}};
\node [numbers] at (32.5, 14.5) {\huge \textbf{6}};
\node [numbers] at (-2.5, 5.5) {\huge \textbf{7}};
\node [numbers] at (8, 5.5) {\huge \textbf{8}};
\node [numbers] at (18.5, 5.5) {\huge \textbf{9}};
\node [numbers, yshift=-.1cm, xshift=.1cm] at (29, 5.5) {\huge \textbf{10}};

\newcommand\txta{14.28}
\newcommand\txtb{5.31}

\node [] at (-0.4, \txta) {\huge \textbf {Expand}};
\node [yshift=0.05cm] at (6.2, \txta) {\huge \textbf {Score}};
\node [yshift=0.05cm] at (13.2, \txta) {\huge \textbf{ Select}};
\node [] at (20.6, \txta) {\huge \textbf {Expand}};
\node [yshift=0.05cm] at (27.2, \txta) {\huge \textbf {Score}};
\node [yshift=0.05cm] at (34.3, \txta) {\huge \textbf {Select}};

\node [] at (-0.3, \txtb) {\huge \textbf {Expand}};
\node [] at (9.8, \txtb) {\huge \textbf {Score}};
\node [] at (20.3, \txtb) {\huge \textbf {Select}};
\node [] at (30.4, \txtb) {\huge \textbf{$\surd$}};

\end{tikzpicture}

%% file: tex/experiments.tex
\begin{table*}
\begin{center}
%    \scalebox{0.9}{
      \input{tables/merged_intrinsic}
%    }
%    \vspace{-2mm}
\end{center}
    \caption{Comparison of static and contextualized scorers in our rule synthesis approach on a held-out portion of our UMBC synthetic data. All methods were allowed to search until they reached the maximum number of explored states.  Shown here are the number of rules successfully found, how many steps were required, and the ceiling (i.e., minimum steps possible using an Oracle). These statistics are averages over 1000 rules of different lengths.}
  \label{tab:intrinsic_merged}
% \vspace{-5mm}
\end{table*}

\section{Experiments and Results}

We evaluate our system both intrinsically and extrinsically.  
The \textit{intrinsic} evaluation is  to determine whether or not the contextualized model 
reduces the number of search steps needed to find a valid rule.  
The \textit{extrinsic} evaluation applies our rule synthesis approach to an information extraction task.

\subsection{Intrinsic Evaluation}
For the intrinsic evaluation, we compare how quickly a valid rule can be found when search is guided by our contextualized scorer versus the static scorers,
measured on a held-out portion of our randomly generated dataset.
We observe from Table \ref{tab:intrinsic_merged} that the transformer-based 
contextualized approach finds {\em more} solutions in {\em fewer} steps. 
This demonstrates that the contextualized scorer is helpful for 
guiding the exploration of the rule search space.

\begin{table*}
% \hskip-5mm
%\scalebox{0.9}{
\begin{center}
\input{tables/few_shot_tacred}

\end{center}
%}
    % \vspace{-2mm}
    \caption{Results of our rule synthesis approach on the testing partition of Few Shot TACRED (micro F1 scores over target relations), compared with a baseline and previous supervised approaches.
    % \todo{is the missing value above on purpose? should we explain it?}}
    }
    % In Goldberg's paper they mentioned they couldn't run sentence-pair in 5-way 5-shot setting due to memory constraints
  \label{tab:tacred}
%  \vspace{-4.5mm}
\end{table*}

\begin{table*}[ht]
  \footnotesize
  \input{tables/examples_fewshot}
      \caption{\footnotesize{Examples of our synthesized rules from the train partition of the Few-Shot TACRED. The relations for the three selected examples are: \texttt{org:city\_of\_headquarters}, \texttt{per:title}, and \texttt{org:subsidiaries}, respectively.} } 
      % Additionally, we added 7 randomly sampled  in Appendix \ref{sec:appendix_random_examples}}}
    \label{tab:examples}
  \end{table*}

\subsection{Extrinsic Evaluation}
To evaluate our approach extrinsically, we want to know how well it performs on an information extraction task.
Ideally, we would evaluate our rule synthesis approach as it is intended to be deployed --- 
with users providing specifications, and on large-scale information extraction projects.
However, that is beyond the scope of the current, initial effort.  
A close proxy is few-shot relation extraction (RE), where a trained system receives a few (often 1 or 5) supporting sentences
for a given relation and is then asked to recognize that relation in an unlabeled query sentence.\footnote{A notable difference between a human evaluation and the few-shot setup is that the former is interactive.  That is, humans could refine the specification based on the results of the previous synthesis, but this is not straightforwardly done in the few-shot setting.}
Recently, \citelanguageresource{Sabo2021RevisitingFR} created a few-shot variant of TACRED \citelanguageresource{zhang2017tacred},\footnote{\url{https://nlp.stanford.edu/projects/tacred/}} a RE 
dataset with 42 possible labels, including \texttt{no\_relation}. %.\footnote{41 relation labels and a \texttt{no\_relation} label.} 
Importantly, in this few-shot variant, the distribution of positives versus negatives was intentionally aligned with that of the real world.

On this task, we compare our rule synthesis approach with one strong baseline and several supervised approaches from previous work (i.e., tuned on the disjoint background set).
These results are provided in Table \ref{tab:tacred}.

\paragraph{\bf Baseline:}
At inference, each query sentence and each of the support sentences contain the \textit{type} of the entities involved in the relation (e.g., \texttt{ORG}, \texttt{PER}, etc).  Using this, we establish a smart random baseline.  Specifically,
the model randomly selects from relations whose supporting sentences have the same entity types, in the same order, as the query sentence, weighted by the number of supporting sentences with that relation.
If there are none, we return \texttt{no\_relation}. 
Additionally, for this baseline we make use of the disjoint background set.  
If there are sentences in that set with the same entity pair as the query, 
we choose one and add it to the supporting sentences with the label \texttt{no\_relation}, available for random selection.

\paragraph{Previous supervised methods:}
We also compare our approach with the current, supervised state of the art (SOA) approaches:
    {\flushleft{{\bf Sentence-Pair} \cite{Gao2019FewRel2T}}}: Concatenates the query sentence with each support sentence and runs the BERT sequence classification model over the concatenated text. If multiple support sentences are available per relation (e.g., in the 5-shot scenario), the score for a relation is obtained as the average over the scores for each sentence. 
    %The prediction is then the relation in which the system is most confident. If the confidence is smaller than $0.5$ for each relation, the prediction is \texttt{no\_relation}}} % ms: removed for space
    {\flushleft{{\bf Threshold} \cite{Sabo2021RevisitingFR}}}: Assigns the \texttt{no\_relation} class to query sentences if the similarity with the support sentences is below a threshold. Otherwise, it assigns the relation with the highest score, as in \textit{Sentence-Pair}. %The threshold is learned using the background training set. 
    {\flushleft{{\bf NAV} \cite{Sabo2021RevisitingFR}}}: 
    A transformer-based relation classifier which uses the background training set to learn a vector for the \texttt{no\_relation} class. At test time, the system computes the similarity between the query sentence and this learned vector, which represents the score for the \texttt{no\_relation} class. The scores for the other relations are obtained using the BERT sequence classification model over the $(\text{query sentence}, \text{support sentence})$ pairs. %With this approach, the \texttt{no\_relation} label is treated as any other label. The prediction is the relation with the highest score.
    {\flushleft{{\bf MNAV} \cite{Sabo2021RevisitingFR}}}: Conceptually similar to NAV, but learns multiple vectors for the \texttt{no\_relation} class. %According to the authors, using multiple vectors associated with the \texttt{no\_relation} label eases the embedding space constraint. Intuitively, the concept of \texttt{no\_relation} is much more broad than that of a single relation, as it means that between the given entities there is a different relation than the ones available in the support set, or there is no relation altogether.

\subsubsection{Extrinsic results} 
First, we note that our proposed baseline performs well, outperforming all of the more expensive BERT-based models on the harder 5-way 1-shot setting. 
Second, our proposed method surpasses the previous state-of-the-art method on the 5-way 1-shot setting, while obtaining competitive performance in the 5-way 5-shot setting. Besides the higher performance in the more challenging 5-shot 1-way setting, note that the output of our approach is a set of \textit{human-interpretable rules}, while the outputs of the other approaches are statistical models that produce the final label without explaining their decisions. 
In other words, previous work is much more opaque, and thus more difficult to interpret, debug, adjust, maintain, and
% Though not often discussed, this is critical when considering the viability of an approach in a real-world setting. 
% \todo{something about hidden biases?}
protect from hidden biases present in the training data \cite{kurita2019measuring,sheng2019woman}.

\subsubsection{Synthesized rules}
We give examples of specifications and the corresponding synthesized rule in Table \ref{tab:examples}.
% , if such a rule was found. 
For space, we list only the highlighted spans. %, not the complete sentence. 
Notably, the system generalizes at different levels, depending on the data available. 
In longer surface rules our model prefers part-of-speech tag constraints, which helps generalization. In rules over simplified syntax, which are often shorter, our model tends to choose lemma or word constraints.
% \todo{needs to be expanded, with specific examples from the table}

% We observed the system with the contextual weights
% that the systems tends to favor part-of-speech tag constraints, which tend to generalize better. We think that this is a by-product of our optimization procedure. Because we predict the score of a transition per sentence, but the final rule should be valid for the complete specification (with possibly multiple sentences), the system tends to favor part-of-speech tag constraints. Indeed, we observed that the our system with contextual weights uses part-of-speech tags constraints in \todo{X\%}, while the static weights approach uses part-of-speech tags constraints only in \todo{Y\%}. 
% Additionally, we list 7 randomly sampled clusters together with their synthesized rule in the Appendix \ref{sec:appendix_random_examples}. 

% BS: removed for space
% \todo{either add an example like this or remove}
% Further, we observe that when the system fails to find a rule, the semantics of the highlighted spans across sentences differ significantly. 
% We suspect that this is a limitation of our clustering technique, as the average embedding of a span is not expressive 
% enough.

Overall, our results indicate that our approach provides a good compromise between the interpretability of 
hand-made rules and the performance of more opaque neural methods.
%provides a feasible approach to automatically generating rules and is a direction worth exploring.

%% file: tables/merged_intrinsic.tex
%{\footnotesize
% \usepackage{multirow}
\begin{tabular}{lrrr}
    \toprule
% \multicolumn{1}{c|}{\multirow{2}{*}{Execution}} &
  % \multicolumn{2}{c|}{With Reward} &
  % \multicolumn{2}{c}{Without Reward} \\
\multicolumn{1}{c}{} &
  \multicolumn{1}{c}{\begin{tabular}[c]{@{}c@{}}Static\\ Weights\end{tabular}} &
  % \multicolumn{1}{c|}{\begin{tabular}[c]{@{}c@{}}Contextual\\ Weight\end{tabular}} &
  \multicolumn{1}{c}{\begin{tabular}[c]{@{}c@{}}Static\\ + Score Aug.\end{tabular}} &
  \multicolumn{1}{c}{\begin{tabular}[c]{@{}c@{}}Contextual\\ Weights\end{tabular}} \\ 
      \midrule
  % \multicolumn{4}{c}{UMBC} \\ 
  %    \midrule      
    Timeout        & 1k states  & 1k states & 1k states \\
    Rules found     & 283/1000 & 581/1000  & 862/1000  \\
    Ceiling avg    & 12.2      & 12.2      & 18.7      \\
    Ceiling median & 14.0      & 14.0      & 17.0      \\
    Ceiling max    & 24        & 24        & 42        \\
    Ceiling min    & 9         & 9         & 9         \\
    Steps avg      & 432.3      & 98.1  & 55.9      \\
    Steps median   & 365.0      & 48.5  & 24.0      \\
    Steps max      & 999        & 676   & 845   \\
    Steps min       & 87         & 13   & 9         \\
    % \midrule
    %  \multicolumn{4}{c}{TACRED} \\ 
    %   \midrule      
    % Timeout         & 1k states  & 1k states      & 1k states \\
    % % Rules found     & 676/3928   & 686/3928       & 1136/3948 \\ % Without trivial cases
    % Rules found     & 971/3948   & 981/3948       & 1431/3948 \\
    % Steps avg       & 203        & 141            & 144       \\
    % Steps median    & 73         & 36             & 34      \\
    % Steps max       & 853        & 929            & 995      \\
    % Steps min       & 14         & 12             & 10      \\ 
    \bottomrule
\end{tabular}
%}

%% file: tables/few_shot_tacred.tex
\begin{tabular}{l|rr}
model         & 5-way 1-shot     & 5-way 5-shot     \\ \hline
Baseline      & 10.82  $\pm$ 0.01\% & 10.90 $\pm$ 0.01\% \\
\hline
Sentence-Pair & 10.19 $\pm$ 0.81\%  & -                \\
Threshold      & 6.87 $\pm$ 0.48\%   & 13.57 $\pm$ 0.46\%  \\
NAV           & 8.38 $\pm$ 0.80\%    & 18.38 $\pm$ 2.01\%  \\
MNAV          & 12.39 $\pm$ 1.01\%  & \textbf{30.04 $\pm$ 1.92\%}  \\
\hline
\textbf{Ours}          & \textbf{15.40 $\pm$ 1.21\%}  & 24.16 $\pm$ 0.44\%     
\end{tabular}

%% file: tables/examples_fewshot.tex
{\footnotesize
\begin{tabular}{l}
\toprule
\textbf{Span} \\
~~~~~~~~ORGANIZATION , which is based in CITY\\
\textbf{Synthesized surface rule:} \\
~~~~~~~~\texttt{[entity=organization] [tag=","] [tag=WDT] [tag=VBZ] [tag=VBN] [tag=IN] [entity=city]} \\
\textbf{Synthesized simplified syntax rule:} \\
~~~~~~~~\texttt{[entity=organization] [tag=NN] [word=based] [entity=city]} \\
\midrule
\textbf{Span} \\
~~~~~~~~PERSON was an indefatigable and enthusiastically received TITLE \\
\textbf{Synthesized surface rule:} \\
~~~~~~~~\texttt{[word=person] [lemma=be] [tag=DT] [tag=JJ] [word=and] [tag=RB] [tag=VBD] [word=title]} \\
\textbf{Synthesized simplified syntax rule:} \\
~~~~~~~~\texttt{[entity=person] [word=indefatigable] [entity=title]} \\
\midrule
\textbf{Span} \\
~~~~~~~~ORGANIZATION , which represents ORGANIZATION \\
\textbf{Synthesized surface rule:} \\
~~~~~~~~\texttt{[entity=organization] [tag=","] [word=which] [tag=VBZ] [entity=organization]} \\
\textbf{Synthesized simplified syntax rule:} \\
~~~~~~~~\texttt{[entity=organization] [lemma=represent] [entity=organization]} \\
% \midrule
% \textbf{Spans in cluster:} \\
% ~~~~~~~~died of a \\
% \textbf{Synthesized rule:} \\
% ~~~~\texttt{[lemma=die] [tag=IN] [tag=DT]} \\
\bottomrule
\end{tabular}
}

%% file: tex/conclusion.tex
\section{Conclusion}
This paper is the first that proposed a synthesis algorithms for rule acquisition. 
Given the importance of explainability in deep learning, we feel this paper opens a new direction in information extraction (IE) that deserves more attention.

% Here we presented initial work on a system for automatically synthesizing rules from a user-provided specification.  
The proposed approach synthesizes rules from a user-provided specification.  
Given one or more (sentence, selection) pairs, the system is able to perform enumerative search 
to find a rule that successfully matches the requested information.
%  and nothing else. % ms: we don't know that it matches nothing else...
Further, we demonstrated that we can utilize the context of the specification to improve the speed of the search, 
with self-supervised pretraining on an automatically generated, generic dataset.
In an extrinsic evaluation that is modeled after the real-world scenario of a user needing to extract a novel relation with only a handful of annotations (i.e., few-shot relation extraction), we showed that our approach outperforms a state-of-the-art BERT model in the 1-shot scenario, and performs competitively in the 5-shot configuration.

% Further, our approach is designed to be compatible with a user interface, such as the one we developed shown in Figure \todo{ref}.
Further, our approach is well-suited to a human-in-the-loop setting, where
a user can select the desired information (or extraction) and request a pattern to be generated, without concern for the underlying representation.
Given our approach, a user could request an alternative rule with the existing specification or augment it to clarify their needs.
Importantly for many settings where systems are deployed long-term and information extraction needs may shift or expand, the resulting rules are able to be understood, modified, and maintained by human users.
% Crucially, the user can request an alternative rule to be synthesized if they do not like the returned rule
%(this can be informed by applying the proposed rule to a corpus and inspecting the results).
%Additionally, the user can augment the specification, adding additional sentences with or without highlighted selections, to further guide the
%synthesis process.
%\todo{Becky: we do not discuss this at well in the paper... it feels disconnected. maybe phrase it as future work?}

% In future work, 
% we will expand our domain-specific language for rules to handle syntactic traversals as well. We will also support % ms: we do (some) syntax now, so this is at least partly obsolete
% we will work on generating several rules (i.e., a grammar) that \textit{together} match the specification, rather than
% requiring a single rule to meet all needs.
% \todo{more and better...!}